\documentclass[10pt,twocolumn,letterpaper]{article}

\usepackage{iccv}
\usepackage{times}
\usepackage{epsfig}
\usepackage{graphicx}
\usepackage{amsmath}
\usepackage{amssymb}
\usepackage{bm}
\usepackage{subfigure}
\usepackage{booktabs}
\usepackage{color}
\usepackage{authblk}
\usepackage[dvipsnames]{xcolor}
\usepackage[pagebackref=ture,pdftex,colorlinks,citecolor=blue,bookmarks=false]{hyperref}
\usepackage[numbers,sort]{natbib}
\newcommand*{\affaddr}[1]{#1} 
\newcommand*{\affmark}[1][*]{\textsuperscript{#1}}
\newcommand*{\email}[1]{\texttt{#1}}

\iccvfinalcopy 

\def\iccvPaperID{331} 

\ificcvfinal\fi
\begin{document}

\title{Recurrent Scale Approximation for Object Detection in CNN}

\author{%
Yu Liu\affmark[1,2], Hongyang Li\affmark[2], Junjie Yan\affmark[1], Fangyin Wei\affmark[1], Xiaogang Wang\affmark[2], Xiaoou Tang\affmark[2]\\
	\vspace{-.2cm}
\affaddr{\affmark[1]SenseTime Group Limited}\\
\affaddr{\affmark[2]Multimedia Laboratory at The Chinese University of Hong Kong}\\
\email{liuyuisanai@gmail.com,\{yangli,xgwang\}@ee.cuhk.edu.hk, \{yanjunjie,weifangyin\}@sensetime.com, xtang@ie.cuhk.edu.hk}
}

\maketitle


\vspace{-0.3cm}
\begin{abstract}
      

Since convolutional neural network (CNN) lacks an inherent mechanism to handle large scale variations, we always need to compute feature maps multiple times for multi-scale object detection, which has the bottleneck of computational cost in practice.
To address this, we devise a recurrent scale approximation (RSA) to compute feature map once only, and only through this map can we approximate the rest maps on other levels. 
At the core of RSA is the recursive rolling out mechanism: given an initial map at a particular scale, it  generates the prediction at a smaller scale that is half the size of input.
To further increase 
efficiency and accuracy, 
we (a): design a scale-forecast network to globally predict potential scales in the image 
since there is no need to compute maps on all levels of the pyramid. 
%
%
(b): propose a landmark retracing network (LRN) to trace back locations of the regressed landmarks and generate a confidence score for each landmark; LRN can effectively alleviate false positives caused by the accumulated error in RSA.
The whole system can be trained end-to-end in a unified CNN framework. 
Experiments demonstrate that our proposed algorithm is superior against state-of-the-art methods on face detection benchmarks and achieves comparable results for generic proposal generation. The source code of our system is available.\footnote{Our codes and annotations mentioned in Sec.4.1 can be accessed at {\color{Rhodamine} \texttt{github.com/sciencefans/RSA-for-object-detection}}}.

\end{abstract}
	\vspace{-.2cm}
\section{Introduction}
\label{intro}
Object detection is one of the most important tasks in computer vision. The convolutional neural network (CNN) based approaches have been widely applied in object detection and recognition with promising performance \cite{fast_rcnn,resNet,faster_rcnn,zeng2016crafting,yu2016poi,liu2017quality,leng20163d, li2017we, liu2017learning}. To localize objects at arbitrary scales and locations in an image, we need to handle the variations caused by appearance, location and scale. Most of the appearance variations can now be handled in CNN, benefiting from the invariance property of convolution and pooling operations. The location variations can be naturally solved via sliding windows, which can be efficiently incorporated into CNN in a fully convolutional manner. However, CNN itself does not have an inherent mechanism to handle the scale variations.

\begin{figure}[t]
	\begin{center}
		\includegraphics[width=0.5\textwidth]{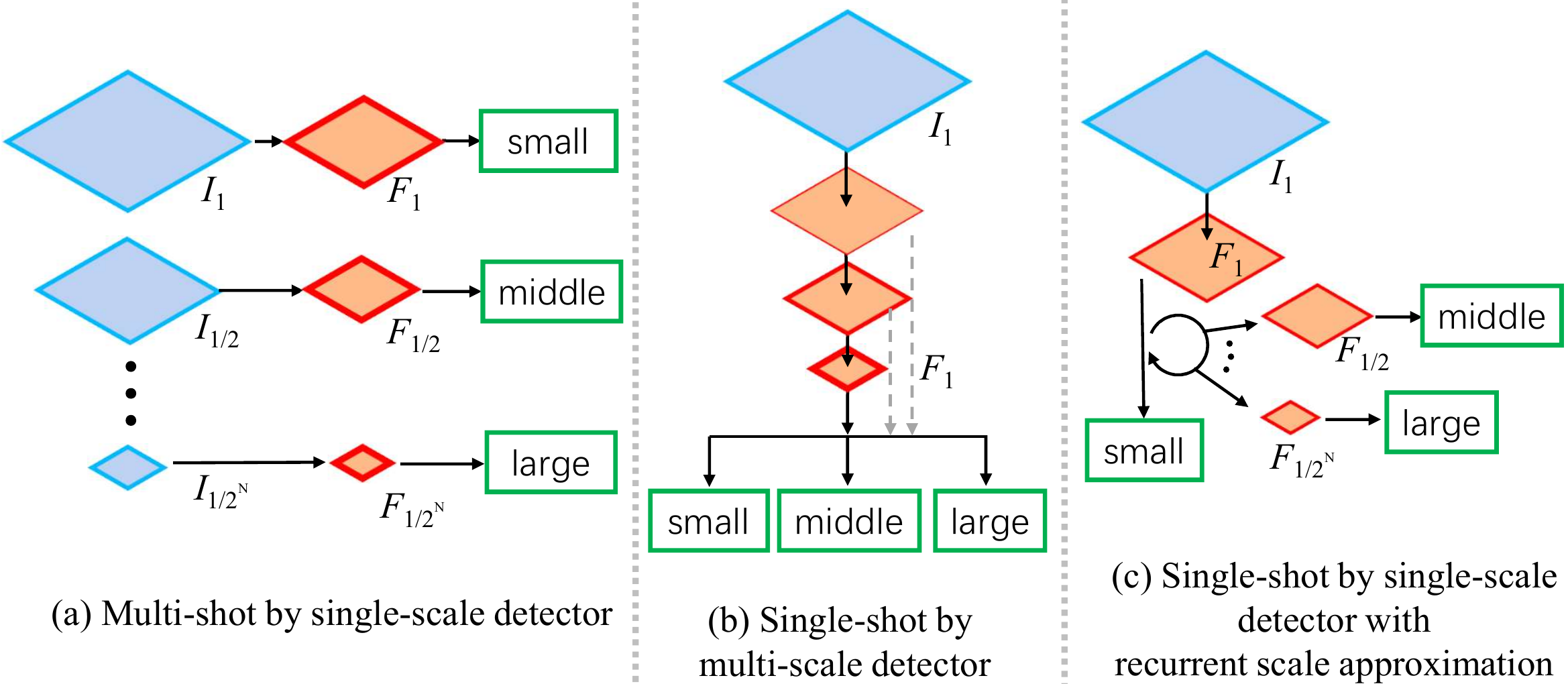}
	\end{center}
	\vspace{-.1cm}
\caption{Different detection pipelines. (a) Image pyramid is generated for multi-scale test. The detector only handles a specific range of scales. (b) Image is forwarded once at one scale and the detector generates all results. (c) Our proposed RSA framework. Image is forwarded once only and feature maps for different scales are approximated by a recurrent unit. Blue plates indicate images of different scales and orange plates with red boarder indicate CNN feature maps at different levels.	}
	\label{fig1}
	\vspace{-.3cm}
\end{figure}
The scale problem is often addressed via two ways, namely, multi-shot by single-scale detector and single-shot by multi-scale detector. The first way, as shown in Fig.~\ref{fig1}(a), handles objects of different scales independently by resizing the input into different scales and then forwarding the resized images multiple times 
for detection \cite{overfeat,li2015convolutional,chen2016supervised}. 
Models in such a philosophy  probably have the highest recall as long as the sampling of scales is dense enough, but they suffer from high computation cost and more false positives. 
The second way, as depicted in Fig.~\ref{fig1}(b), forwards the image only once and then directly regresses objects at multiple scales \cite{faster_rcnn,redmon2016you,liu2016ssd}.
Such a scheme takes the scale variation as a black box. Although more parameters and complex structures would improve the performance, the spirit of direct regression still has limitations in real-time applications, for example in face detection, the size of faces can vary from $20\times 20$ to $1920\times 1080$. 

To handle the scale variation in a CNN-based detection system in terms of both efficiency and accuracy, we are inspired by the fast feature pyramid work proposed by Doll{\'a}r \textit{et al.} ~\cite{dollar2014fast}, where a detection system using hand-crafted features is designed for pedestrian detection. 
It is found that image gradients across scales can be predicted based on natural image statistics. 
They showed that dense feature pyramids can be efficiently constructed on top of coarsely sampled feature pyramids.  
In this paper, we extend the spirit of fast feature pyramid to CNN and go a few steps further. 
Our solution to the feature pyramid in CNN descends from the observations of modern CNN-based detectors, including Faster-RCNN \cite{faster_rcnn}, R-FCN~\cite{NIPS2016_6465}, SSD~\cite{liu2016ssd}, YOLO~\cite{redmon2016you} and STN~\cite{chen2016supervised}, where
 feature maps are first computed and the detection results are decoded from the maps afterwards. 
However, the computation cost of generating  feature maps becomes a bottleneck for methods \cite{overfeat,chen2016supervised} using multi-scale testing and it seems not to be a neat solution to the scale variation problem. 
%

To this end, our philosophy of designing an elegant detection system is that we calculate the feature pyramid 
\textit{once} only, and only through that pyramid can we \textit{approximate} the rest feature pyramids at other scales. The intuition is illustrated  in Fig.~\ref{fig1}(c). 
In this work, we propose a recurrent scale approximation (RSA, see Fig.~\ref{RSA_sample}) unit to achieve the goal aforementioned. The RSA unit is designed to be plugged at some specific depths in a network and to be fed with an initial feature map at the largest scale. The unit convolves the input in a recurrent manner to generate the prediction of the feature map that is half the size of the input. Such a scheme could feed the network with input at one scale only and approximate the rest features at smaller scales through a learnable RSA unit - a balance considering both efficiency and accuracy.

We propose two more schemes to further save the computational budget and improve the detection performance under the RSA framework.
The first is a scale-forecast network 
to globally predict  potential scales for a novel image and we compute feature pyramids for just a certain set of scales based on the prediction. There are only a few scales of objects appearing in the image and hence
most of the feature pyramids correspond to the background, indicating a redundancy if maps on all levels 
are computed.
The second is a landmark retracing network 
that retraces the location 
of the regressed landmarks in the preceding layers and generates a confidence score for each landmark based on the landmark feature set. The final score of identifying a face within an anchor is thereby revised by the LRN network. Such a design alleviates false positives 
caused by the accumulated error in the RSA unit. 

\begin{figure*}[t]
	\begin{center}
		\includegraphics[width=.9\textwidth]{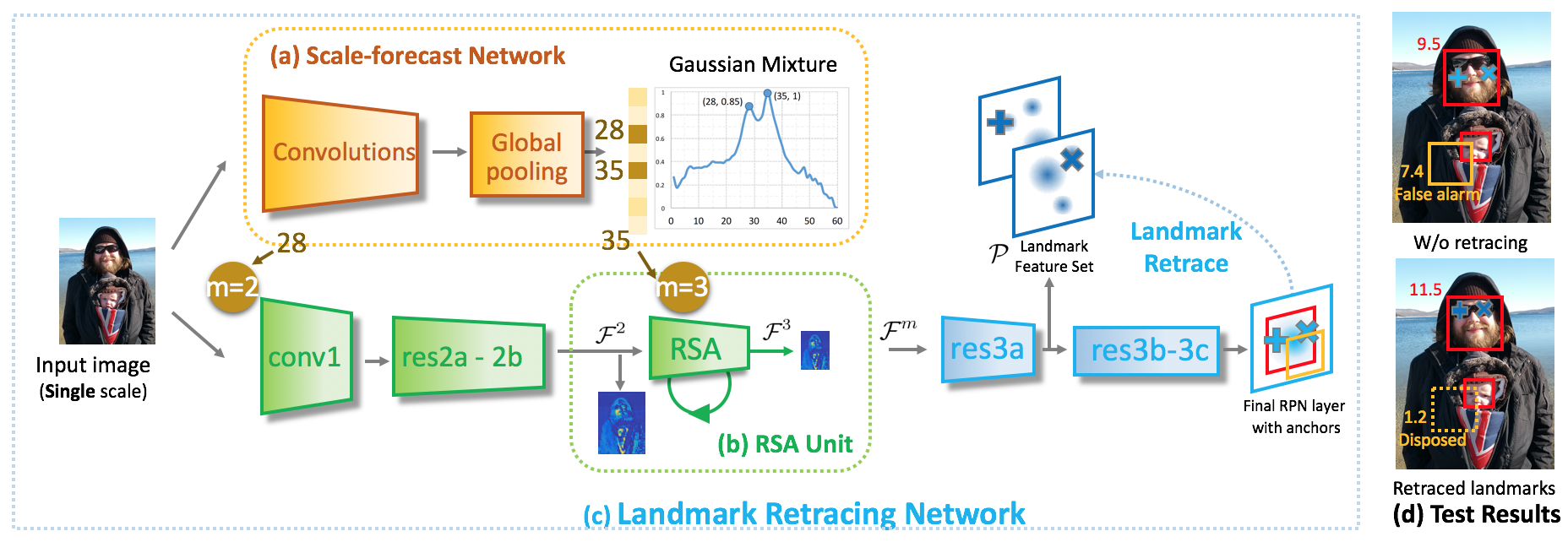}
	\end{center}
	\vspace{-.2cm}
	\caption{Pipeline of our proposed algorithm. (a) Given 
		an image, we predict potential scales from the scale-forecast network and group the results in six main bins ($m=0, \cdots, 5$). (b) RSA unit. The input is resized based on the smallest scale (corresponding to the largest feature map) and the feature maps at other scales are predicted directly from the unit. 
		(c)
		 Given predicted maps, LRN performs landmark detection in an RPN manner. The landmarks can \textit{trace back} locations via regression to generate individual confidence regarding the existence of the landmark. 
		(d) Due to the retracing mechanism, the final score of detecting a face 
		is revised by the confidence of landmarks,
		which can effectively dispose of false positives.
	}
	\label{fig:test_pipe}
	\vspace{-.4cm}
\end{figure*}

The pipeline of our proposed algorithm 
is shown in 
Fig.~\ref{fig:test_pipe}.
The three components 
can be incorporated into a unified CNN framework and trained end-to-end.
Experiments show that our approach is superior to other state-of-the-art methods in face detection and achieves reasonable results for object detection. 

To sum up, our contributions in this work are as follows: 1) We prove that deep CNN features for an image can be approximated from different scales using a portable recurrent unit (RSA), which fully leverages efficiency and accuracy. 2) We propose a scale-forecast network to predict valid scales of the input, which further accelerates the detection pipeline. 3) We devise a landmark retracing network to enhance the accuracy in face detection by utilizing landmark information.

\vspace{-0.3cm}
\section{Related work}
\textbf{Multi-shot by single-scale detector.} 
A single-scale detector detects the target at a typical scale and cannot handle features at other scales. 
An image pyramid is thus formulated and each level in the pyramid is fed  into  the detector.
Such a framework appeared in pre-deep-learning  era \cite{Felzenszwalb2010Cascade,chen2014joint} and usually involves hand-crafted features, \textit{e.g.,}  
HOG~\cite{Dalal2005Histograms} or SIFT~\cite{lowe2004distinctive},
and some classifier like Adaboost~\cite{viola2004robust}, to verify whether the context at each scale contains a target object. 
Recently, 
%
%
some CNN-based methods \cite{li2015convolutional,overfeat} also employ such a spirit to
predict the objectness and class within a sliding window at each scale. 
In this way, the detector only handles features in a certain range of scales and the variance is taken over by the image pyramid, which could reduce the fitting difficulty for detector but potentially increase the computational cost.

\textbf{Single-shot by multi-scale detector.} 
A multi-scale detector takes one shot for the image and generates detection results aross all scales.
RPN \cite{faster_rcnn} and YOLO~\cite{redmon2016you} have fixed size of the input scale, and proposals for all scales are generated in the final layer by using multiple classifiers. 
However,  it is not easy to detect  objects in various scales based on the final feature map.
Liu \textit{et al.} \cite{liu2016ssd} resolved the problem via a multi-level combination of predictions from  feature maps on 
different scales.
And yet it still needs a large model for large receptive field for detection. 
%
Other works \cite{lin2016feature,li2017zoom}
proposed to merge deep and shallow features in a conv/deconv structure and 
to merge boxes for objects from different scales.
These methods are usually faster than the single-scale detector since it only takes one shot for image, but the large-scale invariance  has to be learned by an expensive feature classifier, which is unstable and heavy.

\textbf{Face detection.} Recent years have witnessed a  performance boost in  face detection, which takes 
 advantage of the development in fully convolutional network \cite{farfade2015multi,li2016face,yu2016unitbox,yang2015convolutional}.
%
%
Multi-task RPN is applied \cite{chen2016supervised,ranjan2016hyperface,sun2017face,hao2017scale} 
to generate face confidence and landmarks together. Both single-scale and multi-scale strategies are introduced in these methods.
For example, Chen \textit{et. al} \cite{chen2016supervised} propose a supervised spatial transform layer to utilize landmark information and 
thus enhance the quality of detector by a large margin.  

\section{Our Algorithm}\label{sec:our-algorithm}
In this section, we depict each component of our pipeline (Fig.~\ref{fig:test_pipe}) in detail. 
%
We first devise a scale-forecast network to predict potential scales of the input; 
the RSA unit is proposed to learn and predict  features at smaller scales  based on the output of the scale-forecast network;
the image is fed into the landmark retracing network to detect faces of various sizes, using the scale prediction in Section \ref{sec:scale-forecast-network} and approximation in Section \ref{sec:recurrent-scale-approximation-(rsa)-unit}. 
The landmark retracing network stated in Section \ref{sec:landmark-retracing--network} can trace back features of regressed landmarks and generate individual confidence of each landmark to revise the final score of detecting a face. 
At last, we discuss the superiority of our algorithm's design over other alternatives
in Section \ref{sec:discussion}.

\subsection{Scale-forecast Network}\label{sec:scale-forecast-network}
We propose a scale-forecast network (see Fig.~\ref{fig:test_pipe}(a)) to predict the possible scales of faces given an input image of  fixed size.
The network is a half-channel version of ResNet-18 with a global pooling at the end. The output of this network is a probability vector of $B$ dimensions, where $B=60$ is the predefined number of scales. 
Let $\mathcal{B}=\{0, 1, \cdots, B\}$ denote the scale set, we define the mapping from a face size $x$, in the context of an image being resized to a higher dimension 2048, to the index $b$ in $\mathcal{B}$ as:
\begin{equation}
   b = 10 ( \log_2 x -5).
\end{equation}
For example, if the face has size of 64, its corresponding bin index $b=10$\footnote{Here we assume the minimum and maximum face sizes are 32 and 2048, respectively, if the higher dimension of an image is resized to 2048. In exponential expression, the face size is divided into \textit{six} main bins from $2^5=32$ to $2^{11}=2048$, the denotation of which  will be used later.}.
Prior to being fed into the network, an image is first resized with the higher dimension equal to 224. During training,  the loss of our scale-forecast network is a binary multi-class cross entropy loss:
\begin{equation}
	\mathcal{L}^{SF} = -\frac{1}{B} \sum_{b} p_b \log \hat{p_b} + (1 - p_b) \log (1 - \hat{p_b}),
\end{equation}
where
$p_b, \hat{p}_b$ are the ground truth label and 
prediction of the $b$-th scale, respectively.
Note that the ground truth label for the neighbouring scales $b_i$ of an occurring scale $b*$ ($p_{b*}=1$) is not zero and is defined as the Gaussian sampling score:
\begin{gather}
 p_{b_i}=\texttt{Gaussian}(b_i, \mu, \sigma), b_i \in \mathbb{N}(b*)
\end{gather}
where $\mu, \sigma$ are hyperparameters in the Gaussian distribution and $\mathbb{N(\cdot)}$ denotes the neighbour set. Here we use $ \pm 2$ as the neighbour size and set $\mu$, $\sigma$ to $b*$, $1/\sqrt{2\pi}$, respectively. Such a practice could alleviate the difficulty of feature learning in the discrete distribution between occurring scales (1) and non-occurring scales (0). 

For inference, we use the Gaussian mixture model to determine the local maximum and hence the potential occurring scales.  
Given observations $\bm{x}$, the distribution, parameterized by $\bm{\theta}$, can be decomposed into $K$ mixture components:
\begin{equation}
 p(\bm{\theta} | \bm{x}) = \sum_{i=1}^K \phi_i \mathcal{N} (\bm{\mu}_i, \bm{\Sigma}_i),
\end{equation}
where the $i$-th component is characterized by Gaussian distributions with weights $\phi _i$, means $\bm{\mu}_i$ and covariance matrices $\bm{\Sigma}_i$. Here $K=\{1,...,6\}$ denotes selected scale numbers of six main scales from $2^5$ to $2^{11}$ and the scale selection is determined by the threshold $\phi _i$ of each component. Finally the best fitting model with a specific $K$ is used. 
%



\subsection{Recurrent Scale Approximation (RSA) Unit}\label{sec:recurrent-scale-approximation-(rsa)-unit}
The recurrent scale approximation (RSA) unit is devised to predict feature maps at smaller scales given a map at the largest scale. Fig.~\ref{fig:test_pipe} depicts the 
RSA unit.
The network architecture follows a build-up similar to the residual network \cite{resNet}, where we reduce the number of channels in each convolutional layer to half of the original version for time efficiency. The structure details are shown in Section~\ref{sec:setup-and-implementation-details}.
Given an input image $ \mathcal{I}$, $\mathcal{I}^m$ denotes the downsampled result of the image with a ratio of $1/2^m$, where $m\in\{0, \cdots, M\}$ is the downsample level and $M=5$. Note that $\mathcal{I}^0$ is the original image. Therefore, there are  six scales in total, 
corresponding to the six main scale ranges defined in the scale-forecast network (see Section \ref{sec:scale-forecast-network}).
Given an input image $\mathcal{I}^m$, we define the output feature map of layer \texttt{res2b} as:
\begin{gather}
f( \mathcal{I} ^m ) = \mathcal{G} ^ m,
\end{gather}
where $f(\cdot)$ stands for a set of convolutions with a total stride of 8 from the input image to the output map. The set of feature maps $\mathcal{G} ^ m$ at different scales serves as the ground truth supervision of the recurrent unit.

The RSA module $\texttt{RSA}(\cdot)$ takes as input the feature map of the largest scale $\mathcal{G} ^ 0$ at first,  and repeatedly outputs a map with half the size of the input map:
\begin{gather}
	h^{(0)} = \mathcal{F} ^ 0 = \mathcal{G} ^ 0, \nonumber \\
	h^{(m)} = \texttt{RSA} \big(h^{(m-1)} | \bm{w}   \big) = \mathcal{F}^m. \label{rsa}
\end{gather}
where $\mathcal{F}^m$ is the resultant map after being rolled out $m$ times and $\bm{w}$ represents the weights in the RSA unit.
The RSA module has four convolutions with a total stride of 2 (1,2,1,1) and their kernal sizes are (1,3,3,1). The loss is therefore the $l_2$ norm between prediction $\mathcal{F}^m$ and supervision $\mathcal{G} ^ m$ across all scales:
\begin{equation}
   \mathcal{L}^{RSA} = \frac{1}{2M} \sum_{m=1}^{M}  \big\| \mathcal{F}^m - \mathcal{G} ^ m  \big\|^2.
\end{equation}
The gradients in the RSA unit are computed as:
\begin{align}
    \frac{\partial \mathcal{L}^{RSA}}{\partial \bm{w}_{xy}} & = \sum_{m}  \frac{\partial \mathcal{L}^{RSA}}{\partial h^{(m)}} \cdot \frac{\partial h^{(m)}}{\partial \bm{w}_{xy}}, \nonumber\\
    & = \frac{1}{M} \sum_{m}  \big( \mathcal{F}^m - \mathcal{G} ^ m  \big) \cdot \mathcal{F}^{m-1}_{xy}, 
\end{align}
where $x$ and $y$ are spatial  indeces in the feature map\footnote{For brevity of discussion, we ignore the spatial weight sharing of convolution here. Note that the weight update in $\bm{w}_{xy}$ also includes the loss from the landmark retracing network. }.

The essence behind our RSA unit is to derive a mapping $\texttt{RSA} (\cdot)  \mapsto  f(\cdot)$ to constantly predict smaller-scale features based on the current map instead of feeding the network with inputs of different scales for multiple times. In an informal mathematical expression, we have:
\begin{equation}
\lim_{0 \rightarrow m } \texttt{RSA} \big(h^{(m-1)} \big)  = f( \mathcal{I} ^m ) = \mathcal{G}^m, \nonumber
\end{equation}
to indicate the functionality of RSA: an approximation to $f(\cdot)$ from the input at the largest scale 0 to its desired level $m$.
The computation cost of generating  feature map $\mathcal{F}^m$ 
using RSA is much lower than that of resizing the image and feeding into the network (\textit{i.e.},  $ f( \mathcal{I} ^m )$ through \texttt{conv1} to \texttt{res2b}; see quantitative results in Section \ref{sec:vsbase}).

\begin{figure}[h]
	\begin{center}
		\includegraphics[width=1\linewidth]{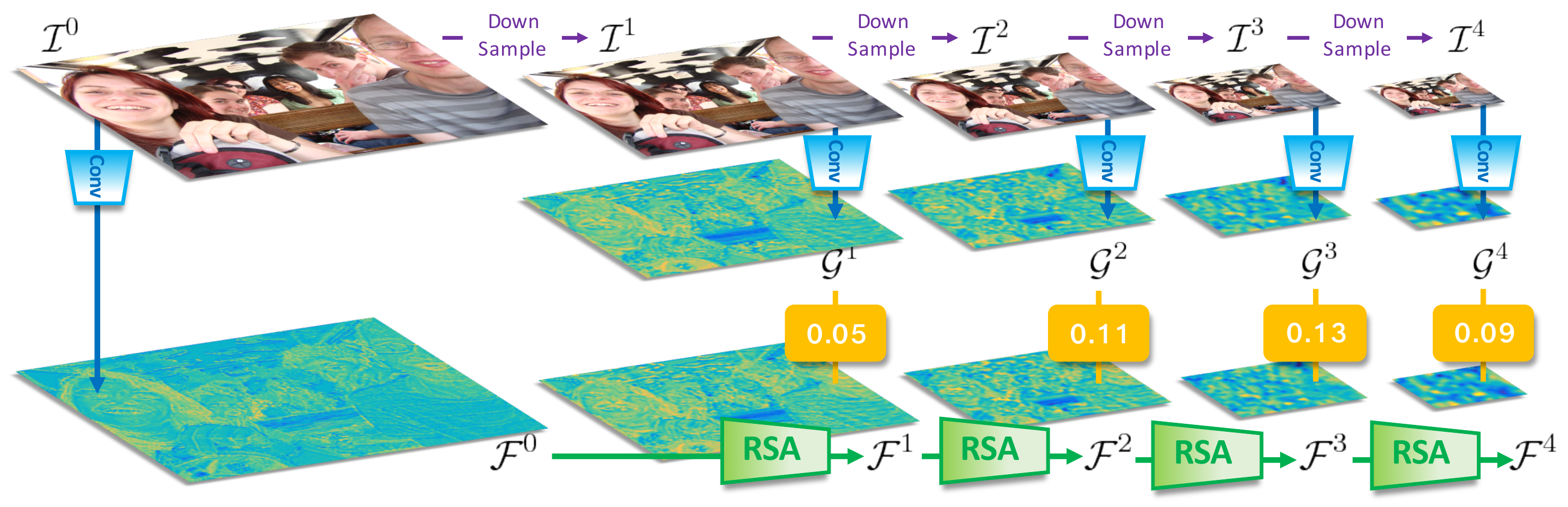}
	\end{center}
	\caption{RSA 
		by rolling out the learned feature map at smaller scales. The number in the orange box indicates the average mean squared error between ground truth and RSA's prediction.}
	\label{RSA_sample}
	\vspace{-.3cm}
\end{figure}

During inference, we first obtain the possible scales of the input from the scale-forecast network.
The image is then resized accordingly so that the smallest scale (corresponding to the largest feature map) is resized to the range of $[64, 128]$. The feature maps at other scales are thereby predicted by the output of RSA unit via Eqn (\ref{rsa}).
Fig.~\ref{RSA_sample} depicts a rolled-out version of RSA to predict feature maps of smaller scales compared with the ground truth. We can observe  from both the error rate and predicted feature maps in each level that RSA is capable of approximating the feature maps at smaller scales.

\subsection{Landmark Retracing  Network}\label{sec:landmark-retracing--network}

In the task of face detection, as illustrated in Fig.~\ref{fig:test_pipe}, the landmark retracing network (LRN) is designed to adjust the confidence  of identifying a face and to dispose of false positives by learning  individual confidence of each regressed landmark. Instead of directly using the ground truth location of landmarks, we formulate such a feature learning of landmarks based on the regression \textit{output} of landmarks in the final RPN layer.

Specifically,
given the feature map $\mathcal{F}$ at a specific scale from RSA ($m$ is dropped for brevity), we first feed it into the \texttt{res3a} layer. There are two branches at the output: 
one is the landmark feature set $\mathcal{P}$ to predict the individual score of each landmark in a spatial context. The number of channels in the set equals to the number of landmarks. Another branch is continuing the standard RPN \cite{faster_rcnn} pipeline (\texttt{res3b-3c}) which generates a set of anchors in the final RPN layer.
Let $\bm{p}_i = [p_{i0}, p_{i1}, \cdots, p_{ik}, \cdots]$ denote the classification probability in the final RPN layer, where $k$ is the class index and $i$ is the spatial location index on the map; $t_{ij}$ denotes the regression target (offset defined in \cite{fast_rcnn}) of the $j$-th landmark in the $i$-th anchor, where $j=\{1, \cdots, 5\}$ is the landmark index. Note that in face detection task, we only have one anchor so that $\bm{p}_i$  contains one element.
In the traditional detection-to-landmark formulation, the following loss, which consists of two heads (\textit{i.e.}, classification and regression), is optimized:
\begin{equation}
\sum_i -\log p_{i k^*}+ \delta(k^*)  \mathcal{S}(\bm{t}_{i} - \bm{t}_{i}^*), \nonumber
\end{equation}
where $\delta(\cdot)$ is the indicator function; $k^*$ denotes the correct label of anchor $i$ and we have only two classes here (0 for background, 1 for positive); $\bm{t}_{i}^*$ is the ground truth regression target and $\mathcal{S}(\cdot)$ is the smoothing $l_1$ loss defined in \cite{fast_rcnn}.

However, as illustrated in Fig.~\ref{fig:test_pipe}(c), using the confidence of anchor $p_{ik^*}$ alone results in 
false positives in some cases, which inspires us to take advantage of the landmark features based on the regression output. The revised classification output, $p_{i k^*}^{trace}(t_{ij})$, now considers both the feature in the final RPN layer as well as those in the landmark feature set:
\begin{align}
p_{i k^*}^{trace}(t_{ij})=\begin{cases} p_{i0}, & k^* = 0, \\  \texttt{max\_pool}(p_{i1}, p_{ij}^{land}), &  k^* = 1,\end{cases} 
\end{align}
where $p_{ij}^{land}$ is the classification output of point $j$ from the landmark feature set $\mathcal{P}$
and it is determined by the regression output:
\begin{equation}
p_{ij}^{land} = \mathcal{P} \big( r(t_{ij}) \big),
\end{equation}
where $r(\cdot)$ stands for a mapping from the regression target to the spatial location on map $\mathcal{P}$. To this end, we have the revised loss for our landmark retracing network:
\begin{equation}
  \mathcal{L}^{LRN} = \sum_i  \bigg[ -\log p_{i k^*}^{trace}(t_{ij}) + \delta(k^*)  \sum_j \mathcal{S}(t_{ij} - t_{ij}^*) \bigg].
\end{equation}
Apart from the detection-to-landmark design as previous work did, our retracing network  also fully leverages
the feature set of landmarks to help rectify the confidence of identifying a face. This is achieved by utilizing the regression output $t_{ij}$ to find the individual score of each landmark on the preceding feature map $\mathcal{P}$. Such a scheme is in a landmark-to-detection spirit.  

Note that the landmark retracing network is trained end-to-end with the RSA unit stated previously. The anchor associated with each location $i$ is a square box of fixed size $64\sqrt{2}$.
The landmark retracing operation is performed only when the anchor is a positive sample. The base landmark location with respect to the anchor is determined by the average location of all faces in the training set.
During test, 
LRN is fed with feature maps at various scales and it treats each scale individually. The final detection result is generated after performing NMS among results from multi-scales.

\subsection{Discussion}\label{sec:discussion}

\textbf{Comparison to RPN.}  The region proposal network \cite{faster_rcnn} takes a set of predefined anchors of different sizes as input and conducts a similar detection pipeline.
Anchors in RPN vary in size to meet the multi-scale training 
constraint. During one iteration of update, it has to feed the whole image of different sizes (scales) from the start to the very end of the network.
In our framework, we resize the image \textit{once} to make sure at least one face 
falls into the size of $[64, 128]$, thus enforcing the network to be trained within a certain range of scales. In this way, we are able to use only one anchor of fixed size. The multi-scale spirit is embedded in an RSA unit, which directly predicts the feature maps at smaller scales. Such a scheme saves parameters significantly and could be considered as a `semi' multi-scale training and `fully' multi-scale test. 
%

\textbf{Prediction-supervised or GT-supervised in landmark feature sets.} 
Another comment on our framework is the supervision knowledge used in training the landmark features $\mathcal{P}$. The features are learned using the prediction output of regression targets $t_{ij}$ instead of the ground truth targets $t_{ij}^*$. In our preliminary experiments, we find that if $p_i^{land}  \sim t_{i}^*$, the activation in the landmark features would be heavily suppressed due to the misleading regression output by $t_{ij}$; however, if we relax the  learning restriction and accept activations within a certain range of misleading locations, \textit{i.e.}, $p_i^{land}  \sim t_{i}$, the performance can be boosted further. Using the prediction of regression as supervision in the landmark feature learning makes sense since: (a) we care about the activation (classification probability) rather than the accurate location of each landmark; (b) $t_i$ and $p_i^{land}$ share similar learning workflow and thus the location of $t_i$ could better match the activation $p_i^{land}$ in $\mathcal{P}$.   

\section{Experiments}

In this section we first conduct the ablation study to verify the effectiveness of each component in our method and compare exhaustively with the baseline RPN \cite{faster_rcnn}; then we compare our algorithm with state-of-the-art methods in face detection and object detection on four popular benchmarks.

\subsection{Setup and Implementation Details}
\label{sec:setup-and-implementation-details}
Annotated Faces in the Wild (AFW) \cite{Zhu2012Face} contains 205 images for evaluating face detectors' performance. 
However, some faces are missing in the annotations and could trigger the issue of false positives, we relabel those missing faces and report the performance difference in both cases.
Face Detection Data Set and Benchmark (FDDB) \cite{fddbTech} has 5,171 annotated faces in 2,845 images. It is larger and more challenging than AFW. 
Multi-Attribute Labelled Faces (MALF) \cite{faceevaluation15} includes 5,250 images with 11,931 annotated faces collected from the Internet. The annotation set is cleaner than that of AFW and it is the largest benchmark for face detection. 

Our training set has 184K images, including 171K images collected from the Internet and 12.9K images from the  training split of Wider Face Dataset \cite{yang2016wider}. All faces are labelled with bounding boxes and five landmarks. The structure of our model is a shallow version of the ResNet \cite{resNet} where the first seven ResNet blocks are used, \textit{i.e.}, from \texttt{conv1} to \texttt{res3c}. We use this model in scale-forecast network and LRN.
All numbers of channels are set to half of the original ResNet model, for the consideration of time efficiency. 
We first train the scale-forecast network and then use the output of predicted scales to launch the RSA unit and LRN. 
%
Note that the whole system (RSA+LRN) is trained end-to-end and the model is trained from scratch without resorting to a pretrained model since the number of channels is halved. 
The ratio of the positive and the negative is $1:1$ in all experiments. 
The batch size is 4; base learning rate is set to 0.001 with a decrease of 6\% every 10,000 iterations. The maximum training iteration is 1,000,000. We use stochastic gradient descent as the optimizer.

\begin{table*}
	\caption{The proposed algorithm is more computationally efficient and accurate by design than baseline RPN.
		Theoretical operations of each component are provided, denoted as `Opts. (VGA input)' below.
		The minimum operation in each component means only the scale-forecast network is used where no face appears in the image; and the maximum operation indicates the amount when faces appear at all scales.
		The actual runtime comparison between ours and baseline RPN is reported in Table \ref{testing_time}.
	}
	\vspace{-.2cm}
	\footnotesize{
		\begin{center}
			\begin{tabular}{ l|c  c c c | c c}
				\toprule
				Component  & Scale-forecast & RSA & LRN  & Total Pipeline & \multicolumn{2}{c}{Baseline RPN}\\
				\midrule
				Structure & tiny ResNet-18 & 4-layer FCN & tiny ResNet-18 & - & single anchor & multi anchors \\
				Opts. (VGA input) & 95.67M & 0 to 182.42M & 0 to 1.3G & \textbf{95.67M to 1.5G} & 1.72G & 1.31G\\
				\midrule
				AP@AFW	& - & - & - & \textbf{99.96\%} & 99.90\% & 98.29\% \\
				Recall@FDDB1\%fpi & - & - & - & \textbf{91.92\%} & 90.61\% & 86.89\% \\
				Recall@MALF1\%fpi & - & - & - & \textbf{90.09\%} & 88.81\% & 84.65\% \\
				\bottomrule
			\end{tabular}
		\end{center}
	}
	\label{network_details}
	\vspace{-.5cm}
\end{table*}

\subsection{Performance of Scale-forecast Network}\label{sec:performance-of-scale-forecast-network}
\vspace{-.2cm}
\begin{figure}[h]
\centering 
\subfigure[AFW]{ 
\includegraphics[width=0.3\columnwidth]{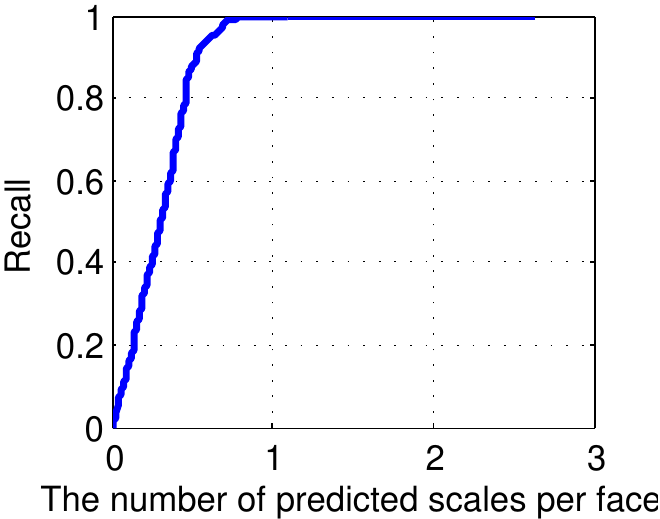}
} 
\subfigure[FDDB]{ 
\includegraphics[width=0.3\columnwidth]{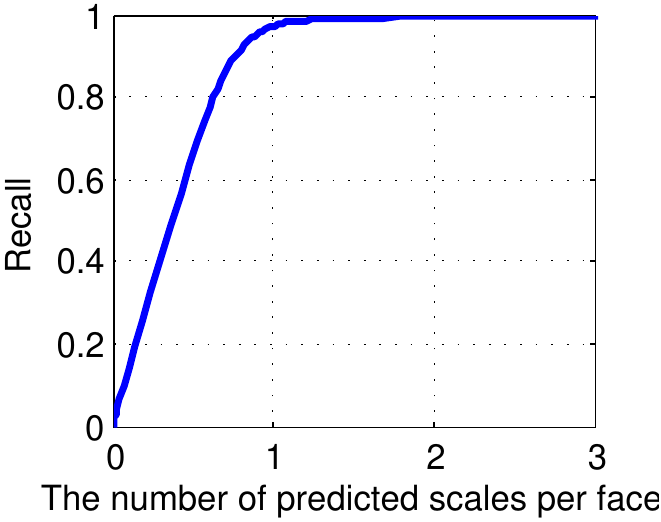}
} 
\subfigure[MALF]{ 
\includegraphics[width=0.3\columnwidth]{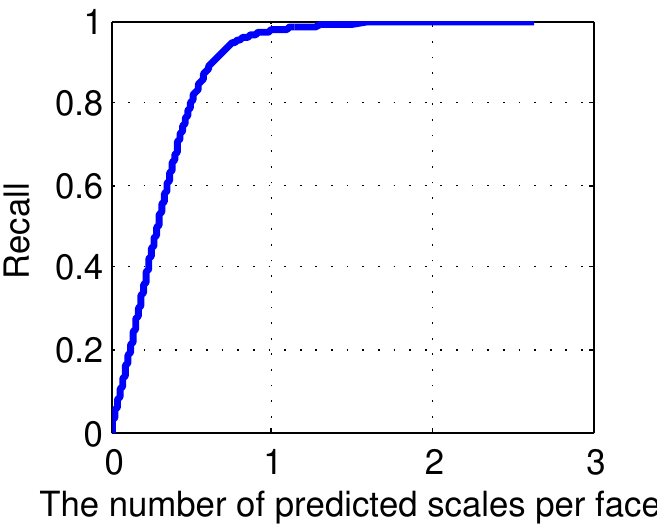}
} 
\caption{Recall v.s. the number of predicted scales per face on three benchmarks. 
	Our scale-forecast network recalls almost all scales when the number of predicted scale per face is 1.} 
\label{scale-forcast_result} 
\end{figure}

The scale-forecast network is of vital importance to the computational cost and accuracy in the networks afterwards.
Fig. \ref{scale-forcast_result} reports the overall recall with different numbers of predicted scales on three benchmarks.
Since the number of faces and the number of potential scales in the image vary across datasets, 
we use  \textit{the number of predicted scales per face} (\textit{x}, total predicted scales over total number of faces) and a global \textit{recall} (\textit{y}, correct predicted scales over all ground truth scales) as the evaluation metric. 
%
We can observe from the results that our trained scale network recalls almost 99\% at $x=1$, 
indicating that on average we only need to generate less than two predictions per image and that we can retrieve all face scales.
Based on this prior knowledge, during inference, we set the threshold for predicting potential scales of the input so that it has approximately two predictions.

\subsection{Ablative Evaluation on RSA Unit}\label{sec:ablative-evaluation-on-rsa-unit}

\begin{figure}[t]
\begin{center}
   \includegraphics[width=.8\linewidth]{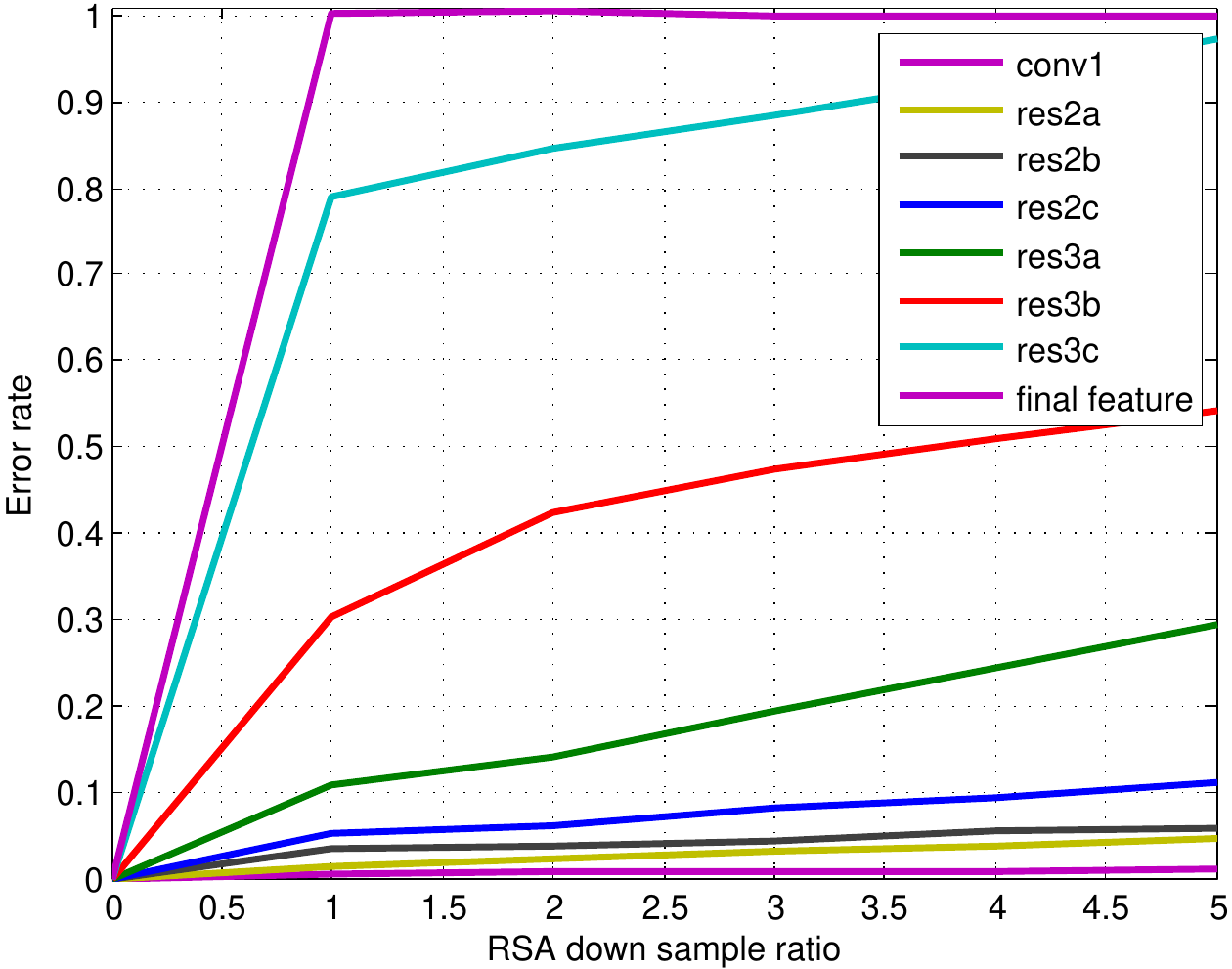}
\end{center}
\vspace{-.4cm}
\caption{Investigation on the source layer to branch out the RSA unit. 
For each case, we report the error rate v.s. the level of down-sampling ratio in the unit.
We can conclude that the deeper the RSA is branched out, the worse the feature approximation at smaller scales will be.
}
\label{featlevel}
\end{figure}

Fig.~\ref{featlevel} investigates the effect of appending the RSA unit to different layers. For each case, the error
rate between the ground truth and corresponding prediction is computed. We define the error rate (ER) on level $m$ as:
\begin{equation}
\texttt{ER}^m = \frac{1}{N} \sum_{i}^{N} \big(  (    \mathcal{F}^{(m)} - \mathcal{G}^{(m)}  ) ./ \mathcal{G}^{(m)} \big)^2,
\vspace{-.2cm}
\end{equation}
where `$./$' implies an element-wise division between maps; $N$ is the total number of samples. 
We use a separate validation set to conduct this experiment. 
The image is first resized to higher dimension being 2048 and the RSA unit predicts six scales defined in Section \ref{sec:scale-forecast-network} (1024, 512, 256, 128 and 64).
Ground truth maps are generated accordingly as we iteratively resize the image (see Fig.~\ref{RSA_sample}).
There are two remarks regarding the result:

First, \textit{feature depth matters.}
Theoretically RSA can handle all scales of features in a deep CNN model and therefore can be branched out at any depth of the network. 
However, results from the figure indicate that as we plug RSA at deeper layers, its performance decades. 
Since features at deeper layers are more sparse and abstract, they barely contain information for RSA to approximate the features at smaller scales. 
For example, in case \texttt{final feature} which means RSA is plugged at the final convolution layer after \texttt{res3c}, the error rate is almost 100\%, indicating RSA's incapability of handling the insufficient information in this layer. 
The error rate decreases in shallower cases.

However, the computation cost of RSA at shallow layers 
is much higher than that at deeper layers, since the stride is smaller and the input map of RSA is thus larger.
The path during one-time forward from image to the input map right before RSA is shorter; and the rolling out time increases accordingly.
Therefore, the trade-off is that we want to plug RSA at shallow layers to ensure a low error rate and at the same time, to save the computational cost.
In practice we choose case \texttt{res2b} to 
be the location where RSA is branched out. Most of the computation happens before layer  \texttt{res2b} and it has an acceptable error rate of 3.44\%. We use this setting throughout the following experiments. 

Second, \textit{butterfly effect exists}. For a particular case, as the times of the recurrent operation increase, the error rate goes up due to the cumulative effect of rolling out the predictions.
For example, in case \texttt{res2b}, the error rate is 3.44\% at level $m=1$ and drops to 5.9\% after rolling out five times. 
Such an increase is within the tolerance of the system and still suffices the task of face detection.
%

%
%
%


\subsection{Our Algorithm vs. Baseline RPN}\label{sec:vsbase}

We compare our model (denoted as RSA+LRN), a combination of the RSA unit and a landmark retracing network,  with the 
region proposal network (RPN) \cite{faster_rcnn}.
%
In the first setting, we use the original RPN with multiple anchors (denoted as RPN\_\texttt{m}) to detect faces of various scales. 
%
In the second setting, we modify the number of anchors to one (denoted as RPN\_\texttt{s}); the anchor can only detect faces in the range from 64 to 128 pixels. 
To capture all faces, it needs to take multiple shots in an image pyramid spirit. 
The network structurse of both baseline RPN and our LRN descend from ResNet-18 \cite{resNet}.
%
Anchor sizes in the first setting RPN\_\texttt{m} are
$32\sqrt{2}, 64\sqrt{2}, \cdots, 1024\sqrt{2}$ 
and they are responsible for detecting faces in the range of $[32,64), [64,128), \cdots, [1024,2048]$, respectively.
In the second setting
RPN\_\texttt{s}, 
we first resize the image length to $64, 256, \cdots, 2048$, then test each scale individually and merge all results through NMS \cite{pami_nms}.

Table~\ref{network_details} shows the theoretical computation cost and test performance of our algorithm compared with baseline RPN. 
We can observe that
RPN\_\texttt{s} needs six shots for the same image during inference and thus the computation cost is much larger than ours or RPN\_\texttt{m};
Moreover, RPN\_\texttt{m} performs worse than the rest two for two reasons:
First,
the receptive field is less than 500 and therefore it cannot see the context of faces larger than 500 pixels; second, it is hard for the 
network (its model capacity much less than the original ResNet \cite{resNet}) to learn the features of faces in a wide scale range from 32 to 2048.
%
%

\begin{table}[t]
	\caption{Test runtime (ms per image) of RSA compared with RPN on three benchmarks. We conduct experiments of each case five times and report the average result
		to avoid system disturbance. }
	\vspace{-.2cm}
\begin{center}
	\footnotesize{
\begin{tabular}{ l|c|c c c}
\toprule
 Speed  & LRN+RSA & LRN & RPN\_\texttt{s} & RPN\_\texttt{m} \\
\midrule
AFW & \textbf{13.95} & 28.84 & 26.85 & 18.92 \\
FDDB & \textbf{11.24} & 27.10 & 25.01 & 18.34 \\
MALF & \textbf{16.38} & 29.73 & 27.37  & 19.37 \\
\midrule
Average & \textbf{14.50} & 28.78 & 26.52 & 18.99 \\
\bottomrule
\end{tabular}
}
\end{center}
\vspace{-.4cm}
\label{testing_time}
\end{table}

\begin{figure*}[ht!]
	\begin{center}
		\includegraphics[width=.97\linewidth]{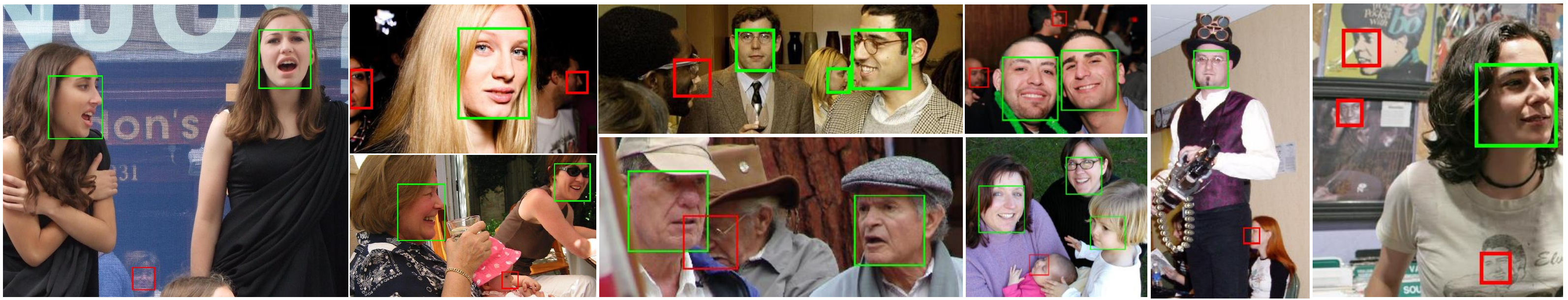}
	\end{center}
	\vspace{-.2cm}
	\caption{Our proposed model can detect faces at various scales, including the green annotations provided in AFW \cite{Zhu2012Face} as well as faces marked in red that are of small sizes and not labeled in the dataset.}
	\label{afw_gt_missed}
\end{figure*}

\begin{figure*}
	\centering 
	\subfigure[FDDB discrete]{ 
		\includegraphics[width=0.49\columnwidth]{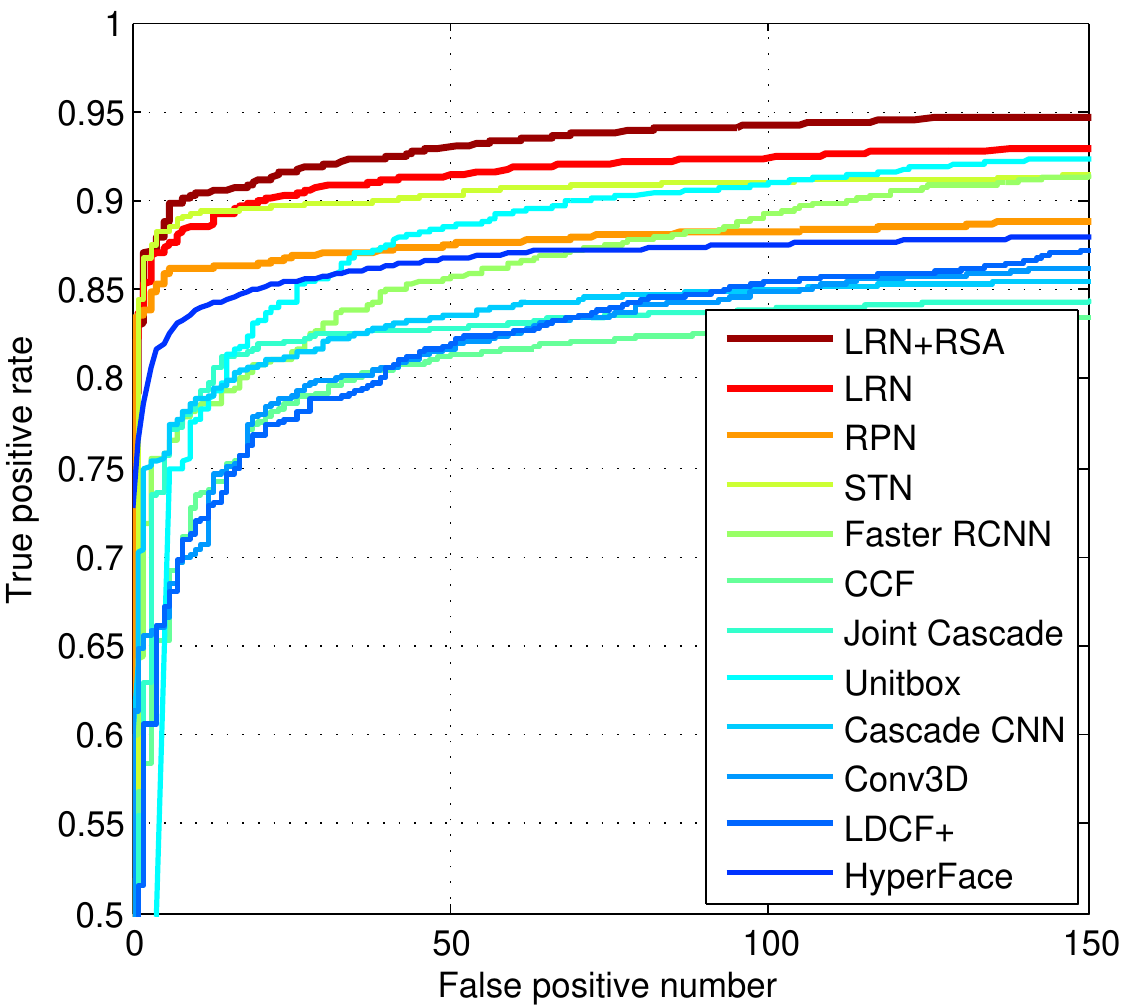}
		\label{fddb_prd}
	} 
	\subfigure[FDDB continuous]{ 
		\includegraphics[width=0.49\columnwidth]{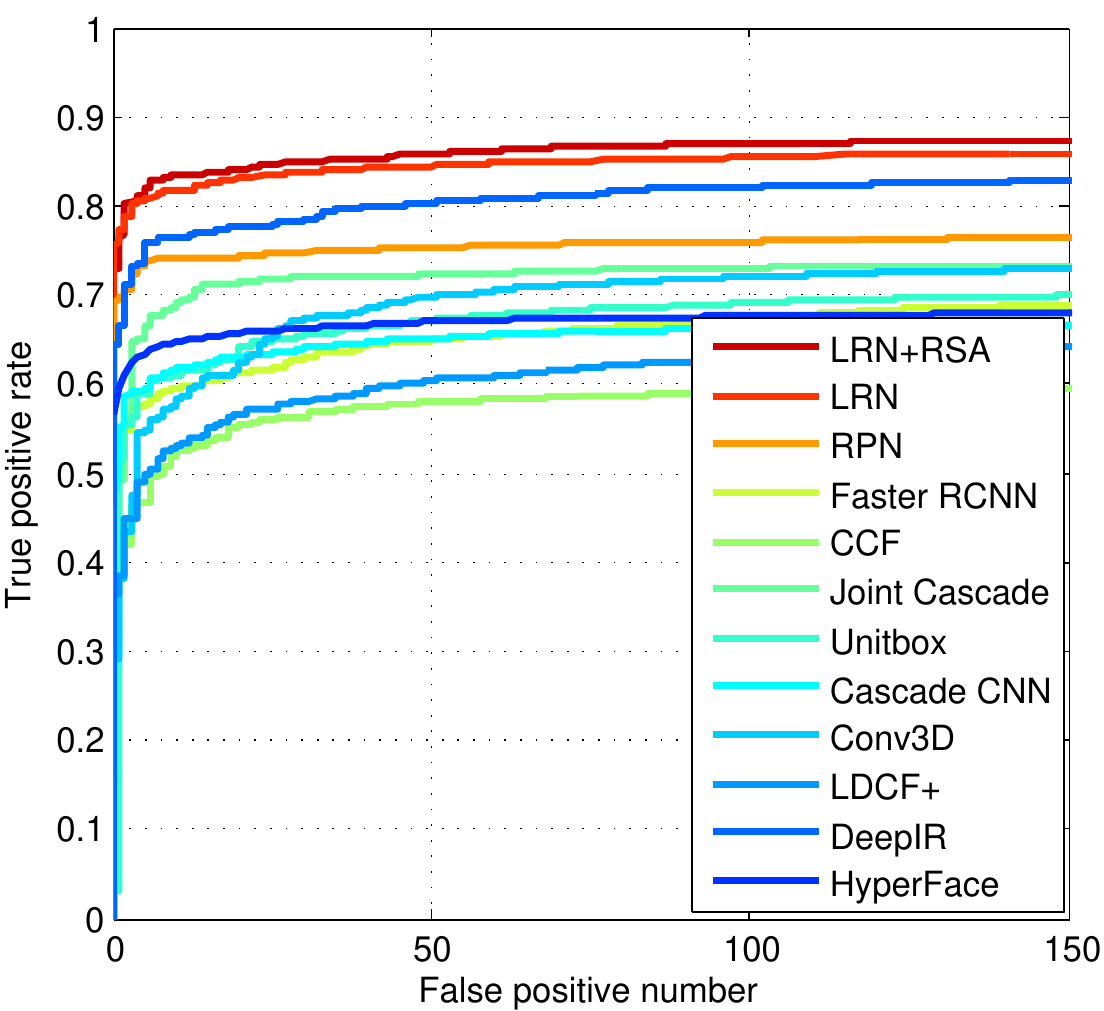}
		\label{fddb_prc}
	} 
	\subfigure[AFW]{ 
		\includegraphics[width=0.49\columnwidth]{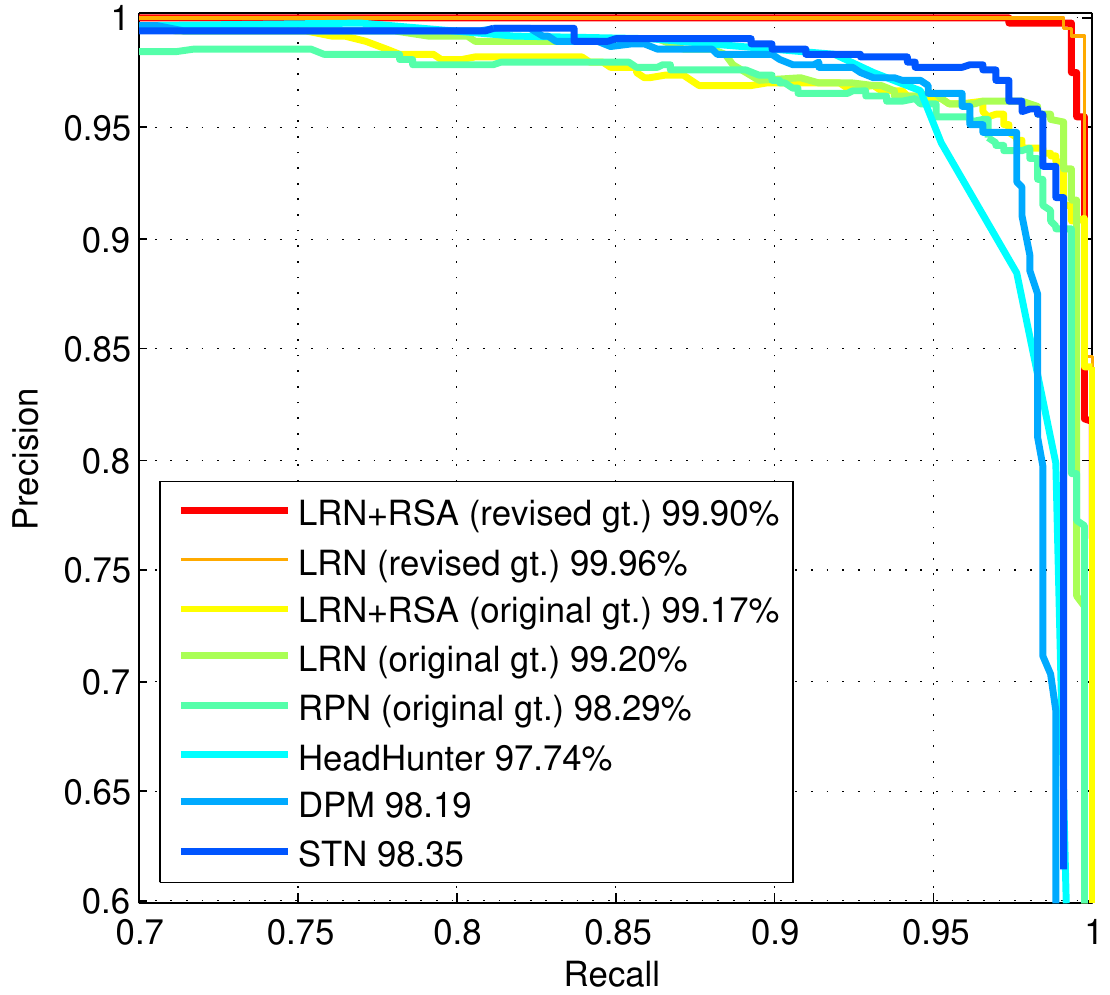}
		\label{afw_pr}
	} 
	\subfigure[MALF]{ 
		\includegraphics[width=0.49\columnwidth]{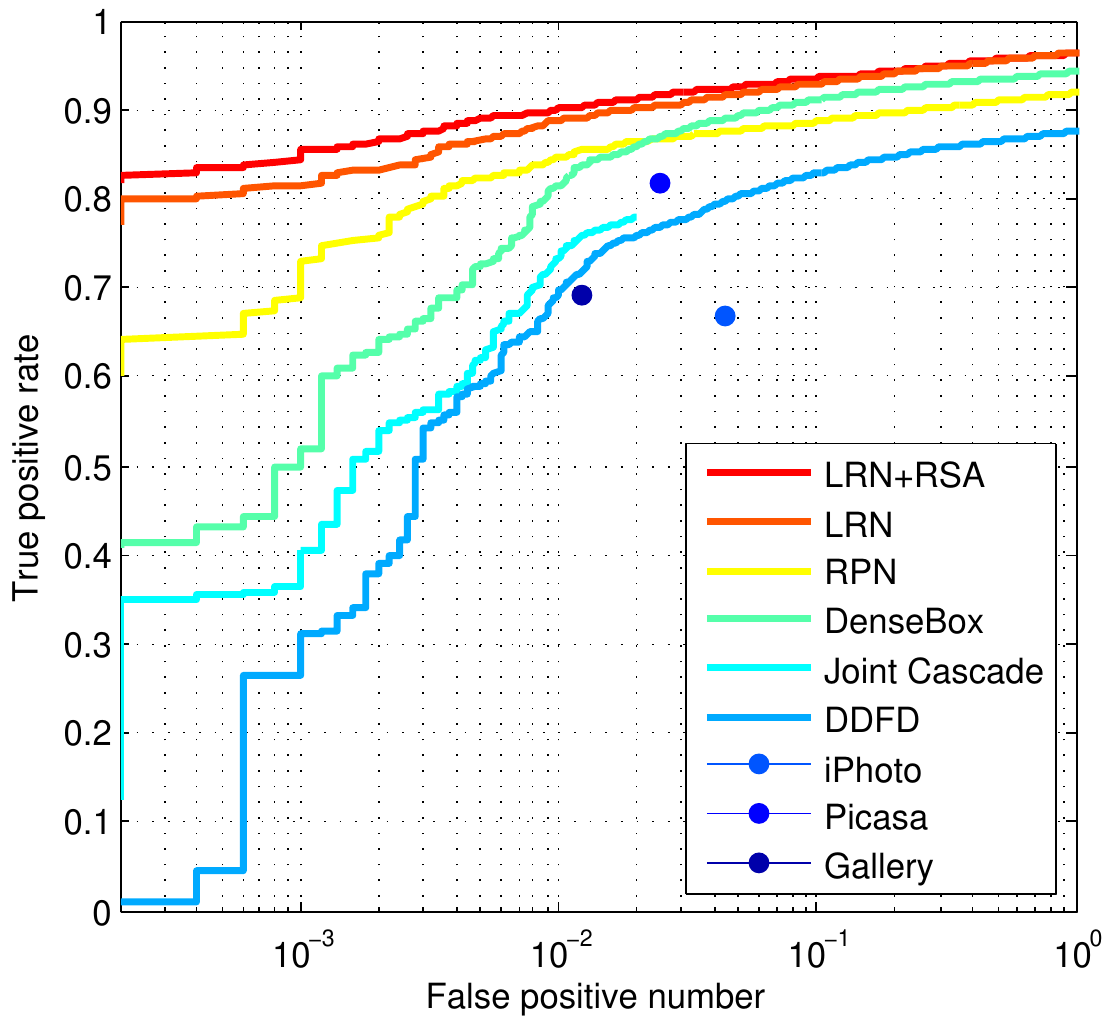}
		\label{malf_pr}
	} 
	\caption{Comparison to state-of-the-art approaches on face detection benchmarks. 
		The proposed algorithm (Scale-forecast network with RSA+LRN, tagged by LRN+RSA) outperforms other methods by a large margin.
		`revised gt.' and 'original gt.' in AFW stand for fully annotated faces by us and partially labeled annotations provided by the dataset, respectively.
	}  
	\label{com_to_soa}
\end{figure*}

Table~\ref{testing_time} depicts the runtime comparison during test. 
The third column LRN means without using the RSA unit.
Our method runs fast enough compared with its counterparts for two reasons. First, there are often one or two valid scales in the image, and the scale-forecast network can automatically select some particular scales, and ignore all the other invalid ones in the multi-scale test stage; 
second, the input of LRN descends from the output of RSA to predict feature maps at smaller scales; it is not necessary to compute feature maps of multiple scales in a multi-shot manner as RPN\_\texttt{m} does.

\subsection{Face Detection}
\label{state_of_the_art}

Fig.~\ref{com_to_soa} shows the comparison against other approaches on three benchmarks. On AFW, our algorithm achieves an AP of 99.17\% using the original annotation and an AP of 99.96\% using the revised  annotation \ref{afw_pr}. On FDDB, RSA+LRN recalls 93.0\% faces with 50 false positives \ref{fddb_prd}. On MALF, our method recalls 82.4\% faces with zero false positive \ref{malf_pr}. 
It should be noticed that the shape and scale definition of bounding box on each benchmark varies. For instance, the annotation on FDDB is ellipse while others are rectangle. To address this, we learn a transformer to fit each annotation from the landmarks. 
This strategy significantly enhances  performance  in the continuous setting on FDDB.

%

\begin{table}
	\caption{Recall (\%) vs. the number of proposals and Speed (ms per image) breakdown on ILSVRC DET \texttt{val2}.}
	\vspace{-.2cm}
	\begin{center}
		\footnotesize{
			\begin{tabular}{l|c c c c |c}
				\toprule
				Recall  		 & 100 	& 300 	& 1000 	& 2000 & Speed\\
				\midrule
				Original RPN 	 & 88.7 & 93.5 & 97.3 & 97.7 & 158 ms \\
				Single-scale RPN & 89.6 & 94.4 & 97.2 & 98.0 & 249 ms \\
				RSA+RPN 		 & 89.1 & 94.4 & 97.2 & 98.0 & 124 ms \\
				\bottomrule
			\end{tabular}
		}
	\end{center}
	\label{imgnt_test}
	\vspace{-.7cm}
\end{table}

\subsection{RSA on Generic Object Proposal}
\label{imagenet}
We now verify that the scale approximation learning by RSA unit also generalizes comparably well on the generic region proposal task.
Region proposal detection is a basic stage for generic object detection task and is more difficult than face detection. \
ILSVRC DET \cite{imagenet_conf} is a challenging dataset for generic object detection. It contains more than 300K images for training and 20K images for validation. 
We use a subset (around 170k images) of the original training set for training, where 
each category has at most 1000 samples; for test we use the \texttt{val2} split \cite{girshick2014rich} with 9917 images.
%
%
We choose the single anchor RPN with ResNet-101 as the baseline. RSA unit is set after \texttt{res3b3}. 
The anchors are of size $128\sqrt{2}$ squared, $128\times256$ and $256\times128$. 
%
%
During training, we randomly select an object and resize the image so that the object is rescaled to
$[128, 256]$. Scale-forecast network is also employed to predict the higher dimension of objects in the image. 

Recalls with different number of proposals are shown in Table~\ref{imgnt_test}.
The original RPN setting has  18 anchors with 3 aspect ratios and 6 scales.
Without loss of recall, RPN+RSA reduces around 61.05\% computation cost compared with the single-scale RPN, when the number of boxes is over 100. RPN+RSA is also more efficient and recalls more objects than original RPN.
Our model and the single-anchor RPN both perform better than the original RPN. 
This observation is in accordance with the conclusion in face detection. 
Overall, our scheme of using RSA plus LRN competes comparably with the standard RPN method in terms of computation efficiency and accuracy.

\section{Conclusion}
In this paper, we prove that deep CNN features of an image can be approximated from a large scale to smaller scales by the proposed RSA unit, which significantly accelerates face detection while achieving comparable results in object detection.
In order to make the detector faster and more accurate, we devise a scale-forecast network to predict the potential object scales.
%
We further design a landmark retracing network to fuse global and local scale information to enhance the predictor. Experimental results show that our algorithm significantly outperforms state-of-the-art methods. Future work includes exploring RSA on generic object detection task. Representation approximation between video frames is also an interesting research avenue.

{\small
\bibliographystyle{ieee}
\bibliography{egbib,dl_iccv17}
}

\end{document}